%% file: main.tex
\documentclass[11pt,twocolumn]{article}
\pdfoutput=1

\usepackage{cite}
\usepackage{amsmath,amssymb,amsfonts}
\usepackage{algorithmic}
\usepackage{graphicx}
\usepackage{textcomp}

\usepackage[utf8]{inputenc}
\usepackage{graphicx}
\usepackage{amsmath}
\usepackage{amssymb}
\usepackage{gensymb}
\usepackage{footmisc}
\usepackage{listings}
\usepackage{url}
\usepackage{footmisc}
\usepackage{makecell}

\usepackage[listofformat=parens, justification=centering, subrefformat=subparens]{subfig}  

\usepackage[top=1.5cm,left=1.2cm,right=1.2cm,bottom=2.5cm]{geometry}

\DeclareGraphicsExtensions{.pdf,.jpg,.png}
\graphicspath{{./graphics/}{./pictures/}{./plot/}}

\usepackage[breaklinks=true,colorlinks,bookmarks=false]{hyperref}

\newcommand\ig[2][1]{\includegraphics[width=#1\textwidth]{#2}}

\newcommand\igp[3][1]{\includegraphics[width=#1\textwidth,page=#2]{#3}}

\renewcommand\sec[1]{Sec.~\ref{sec:#1}}
\newcommand\fig[1]{Fig.~\ref{fig:#1}}
\newcommand\sfig[1]{Fig.~\subref{fig:#1}}
\newcommand\tab[1]{Tab.\;\ref{tab:#1}}

\newcommand\chapter{\section}

\newcommand\Caption[3][]{\caption[#2]{\label{#1}\textsc{#2}. \small\textit{#3}}}

\title{Eight Years of Face Recognition Research:\\ Reproducibility, Achievements and Open Issues}
\author{%
  Tiago de Freitas Pereira,$^1$
  Dominic Schmidli,$^2$
  Yu Linghu,$^2$\\[1ex]
  Xinyi Zhang,$^2$
  S\'ebastien Marcel,$^{1,3}$ and
  Manuel G\"unther$^2$\\[2ex]

  $^1$Idiap Research Institute \quad $^2$University of Zurich \quad $^3$University of Lausanne
}

\date{}

\begin{document}

\maketitle

\begin{abstract}\it
Automatic face recognition is a research area with high popularity.
Many different face recognition algorithms have been proposed in the last thirty years of intensive research in the field.
With the popularity of deep learning and its capability to solve a huge variety of different problems, face recognition researchers have concentrated effort on creating better models under this paradigm.
From the year 2015, state-of-the-art face recognition has been rooted in deep learning models.
Despite the availability of large-scale and diverse datasets for evaluating the performance of face recognition algorithms, many of the modern datasets just combine different factors that influence face recognition, such as face pose, occlusion, illumination, facial expression and image quality.
When algorithms produce errors on these datasets, it is not clear which of the factors has caused this error and, hence, there is no guidance in which direction more research is required.
This work is a followup from our previous works developed in 2014 and eventually published in 2016, showing the impact of various facial aspects on face recognition algorithms.
By comparing the current state-of-the-art with the best systems from the past, we demonstrate that faces under strong occlusions, some types of illumination, and strong expressions are problems mastered by deep learning algorithms, whereas recognition with low-resolution images, extreme pose variations, and open-set recognition is still an open problem.
To show this, we run a sequence of experiments using six different datasets and five different face recognition algorithms in an open-source and reproducible manner.
We provide the source code to run all of our experiments, which is easily extensible so that utilizing your own deep network in our evaluation is just a few minutes away.
\end{abstract}

\input{introduction}
\input{related}
\input{approach}
\input{experiments}
\input{discussion}
\input{conclusion}

\bibliographystyle{IEEEtran}
\bibliography{references}

\end{document}

%% file: introduction.tex
\nocite{gunther2016,gunther2017}

\section{Introduction}
Biometric recognition has attracted much attention in the past decades.
Commonly used examples of biometric recognition include methods of recognizing one's face, iris, voice, ear, palm print, gait, or signature \cite{minaee2021biometrics}.
Face recognition is one of the most popular forms of biometric recognition and its development has made great progress in the last decades, mainly influenced by the availability of different open-source methods for face processing, including face and facial landmark detection and face recognition \cite{wanyonyi2022opensource}.
Furthermore, its field of application is very versatile, as almost every mobile device, including laptops and smartphones, nowadays offers the possibility to unlock its screen through face recognition.
Another popular application is video surveillance and through security cameras \cite{masi2018} where face recognition can help to identify criminals or find missing persons.
In these and many other fields, the need for robust facial recognition systems has increased year over year \cite{guo2019survey} and already in 2007 automatic face recognition has superseded human performance in controlled and constrained environments \cite{otoole2007surpass}.
Thus, in security-relevant applications such as automatic border control, frontal faces, neutral expressions, and good illumination are enforced \cite{delrio2016automated}.
However, such an environment can not always be found.
Especially in outdoor surveillance situations, illumination from the sun is often not ideal for recognizing faces and people may show different expressions and will likely not look into the camera \cite{guenther2017challenge}.
Furthermore, subjects may wear hats or glasses, faces might be partially occluded, and the quality and size of the image can vary greatly \cite{mei2020}.
All these conditions can seriously interfere with the performance of face recognition.
\begin{figure}[t!]
    \centering
    \includegraphics[width=.95\linewidth]{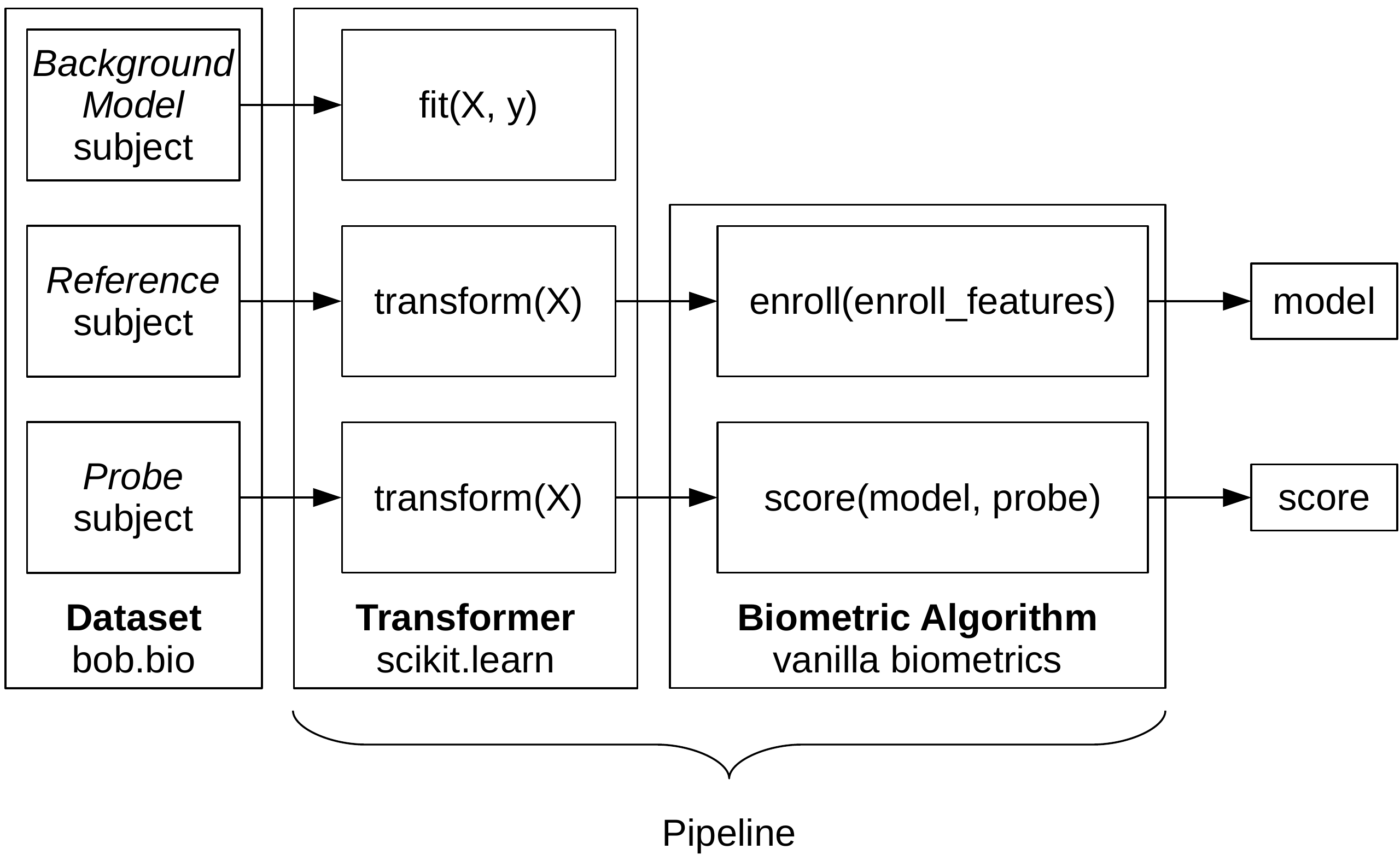}
    \Caption[fig:FR-process]{Biometric Recognition Pipeline in Bob} {
        The dataset implements the evaluation protocol, i.e. which samples to use for enrollment and probing.
        Each subject's data goes through the Transformer (or series of Transformers) to extract features, which are then given to enroll models or to compute similarity scores between model and probe features.
    }
\end{figure}
Before the era of deep learning, face recognition employed hand-crafted features and traditional algorithms.
When conditions for capturing faces are not optimal, such as when there are different facial expressions, lighting conditions or face poses, the performance of these traditional algorithms drop significantly \cite{gunther2016}.
The biggest issue of hand-crafted features is that they fail to capture important information from the faces, and traditional algorithms are not well-suited for all different aspects of face recognition, particularly, for changes in illumination and pose.

This changed with the development of convolutional neural networks, the Compute Unified Device Architecture (CUDA), which allows data-intensive training of deep networks \cite{minaee2021biometrics}, and libraries that easily put all these elements together, such as Theano, Caffe, MxNet, Tensorflow, PyTorch, PaddlePaddle and so on.
Since then, face recognition has been dominated by neural networks, and performance has steadily improved since deep networks automatically learn the best suited features to extract from the images, as well as provide algorithms to learn the different aspects of face recognition in an integrated way.
While researchers first focused on creating deeper networks with the support of advanced network architectures, the attention has shifted toward creating more powerful loss functions that are better adapted to face recognition requirements, i.e., to increase between-class and decrease within-class variability of the deep features extracted from the networks.
One of the methods that has demonstrated a state-of-the-art performance is ArcFace \cite{deng2018arcface}, which uses an additive angular margin loss function to separate the features.
Latest inventions in face recognition include MagFace \cite{meng2021magface}, which include metrics for image quality into training deep networks.

Another main area of research is the creation of datasets, which are not only important for training, but also essential for performance comparison.
Many papers and several surveys \cite{masi2018,guo2019survey,sawant2019age,mei2020,minaee2021biometrics,wanyonyi2022opensource} show the performance of different deep neural networks in challenging environments that achieve impressive results on the LFW \cite{lfw} and IJB-C \cite{ijbc} datasets.
Unfortunately, these datasets include only mixtures of different aspects of face recognition, but it is impossible to evaluate which of these aspects can be considered to be solved and which aspects require more research focus.
Only in rare cases, aspects such as ethnicity \cite{masi2018} or aging \cite{sawant2019age} are evaluated separately.
Also, even though the title of the latest review \cite{wanyonyi2022opensource} suggests to review open-source face recognition \textbf{frameworks}, the authors only investigate different \textbf{parts of the face recognition pipeline} (different datasets, face and facial landmark detection methods, deep learning models and evaluation techniques), but they do not test combinations of different parts since no \textbf{framework} for face recognition is available and maintained -- except for the Bob framework \cite{bob2012,bob2017,guenther2012facereclib} utilized and advertised in this work.
It is important to note that those surveys only duplicate the results promised by the reviewed papers \textbf{without having a chance to reproduce these numbers or to change evaluation criteria}.

The present work aims to close this gap by comparing the performance of state-of-the-art deep neural networks in different challenging face recognition environments in a reproducible, comparable and extensible manner.
To achieve this, four pre-trained networks from the state of the art are examined.
Experiments are performed on six datasets, including AR face \cite{arface}, Multi-PIE \cite{multipie}, SCface \cite{scface}, GBU \cite{gbu}, IJB-C \cite{ijbc}, and MOBIO \cite{mobio}, each of which represents a different aspect of face recognition.
Our implementations of evaluation protocols \cite{gunther2016,gunther2017} allow the isolated consideration of different aspects of face variations, such as different types of occlusion, facial expressions, or poses.
The performed experiments make use of the open-source biometrics recognition pipeline \cite{guenther2012facereclib} of Bob \cite{bob2012,bob2017}.
Together with this study, an open-source implementation to re-run (or at least re-evaluate) the experiments is provided.\footnote{\label{fn:package}\url{https://gitlab.idiap.ch/bob/bob.paper.8years}}

This work extends our previous work that was executed in 2014 and finally published in 2016 \cite{gunther2016} and 2017 \cite{gunther2017}, in which we evaluated the performance of several traditional algorithms on different datasets and, thereby, considered individual challenging face recognition conditions in isolation.
In that study, we found that strong occlusion has a significant impact on recognition rates.
For face recognition across pose, especially with angles deviating beyond $\pm 45\degree$ from frontal, traditional algorithms almost completely fail.
Also, facial image comparison including low-resolution probe images poses a major challenge to those algorithms.

The purpose of the current study is to extend and update our previous open-source package\footnote{\label{fn:FRICE}\url{https://gitlab.idiap.ch/bob/bob.chapter.frice}} in order to see how large the progress of face recognition for various aspects of face recognition has been in the last eight years.
Therefore, we compare the best traditional open-source systems, i.e., the inter-session variability (ISV) modeling \cite{wallace2011intersession,gunther2016} with current state-of-the-art open-source algorithms found today.

The contributions of this work are:
\begin{itemize}
    \item We evaluate current state-of-the-art deep-learning face recognition algorithms on \emph{various aspects of face recognition separately} and compare them with the best traditional face recognition algorithm.
    \item In comparison to our old evaluation, we change to a better-suited evaluation procedure that better \emph{accounts for requirements of real-world applications}.
    \item We show that even extreme facial expressions and large occlusions of the faces  are handled well by deep learning algorithms, but \emph{research should focus more on comparing faces across pose, on low-resolution images} and on open-set identification.
    \item We run all experiments in an \emph{open-source and reproducible manner}, and provide tools to \emph{easily extend this research} to novel future findings and developments.
\end{itemize}

%% file: related.tex
\section{Related Work}

\begin{table}[t]
    \Caption[tab:preprocessing]{Preprocessing}{Summary of the employed eye and mouth positions for the experiments. All parameters are given in Bob's \texttt{(y,x)} order. Right and left eye coordinates were used for frontal faces, eye and mouth for profile faces.}
    \centering\small
    \begin{tabular}{|l||c|c|c|}
        \hline
        & \textbf{Facenet} & \makecell{\textbf{ArcFace-100}, \\ \textbf{Zoo-AttentionNet} \\ and \textbf{Idiap-Resnet50}} & ISV\\\hline\hline
        Resolution & $160\times160$ & $112\times112$ & $80 \times 64$\\\hline\hline
        Right Eye & (32, 39) & (52, 38) & (16, 15)\\\hline
        Left Eye & (32, 120) & (52, 74) & (16, 48)\\\hline\hline
        Eye & (32, 64) & (52, 56) & (16, 25) \\\hline
        Mouth & (106, 64) &(91, 56) & (52, 25) \\\hline
    \end{tabular}

\end{table}

This section provides an overview of the current state of research regarding face recognition and deep learning.
First, available face datasets are described.
Other than a few popular datasets, the focus is put on the datasets utilized in our experiments.
More datasets and algorithms can, for example, be found in \cite{wanyonyi2022opensource}.
The next section gives a brief summary of our previous evaluation \cite{gunther2016,gunther2017}, on which this study is based, and presents the current state of the art in deep learning for face recognition.

\subsection{Datasets}
A major research interest in the area of deep learning lies in the development of new datasets.
There are a large number of facial image datasets that differ greatly in the number of images and identities, as well as in the diversity of the images.
While older datasets have often been split into parts for training and for evaluation of algorithms, data-hungry deep learning methods require more data to train and, therefore, training and evaluation datasets have been disentangled.
This section gives an overview of some commonly used datasets for training and evaluating face recognition models.

\subsubsection{Training Datasets}
The Visual Geometry Group (VGG) in Oxford developed a five-step guide to compile a large dataset, and applied these instructions to images of the celebrities in the Internet Movie Database (IMDB).
The VGGFace dataset \cite{vggface} consists of over 2.6 million images from 2'622 different celebrities, with about five percent of these images showing profile faces, the others being mostly frontal.
The extended VGGFace2 \cite{vggface2} dataset consists about three million images of 9'131 identities with facial images varying in pose, background, age and illumination, yet all images are of relatively high resolution.
%

%
CASIA-WebFace \cite{yi2014} features about half a million images from 10'000 identities.
It is also often used for face verification and face identification.
The images were collected from celebrities of various years of birth.
MS-Celeb-1M (MS1M) \cite{guo2016msceleb1m} is a dataset with ten million images of celebrities collected from the Internet representing a variety of nationalities and professions such as politicians, actors, writers, and singers.
It consists of 100'000 identities in total with about 100 images per identity, with over three quarters of the subjects being female.
Several researchers released extensions of this dataset, most of them handling some mislabeling issues that a dataset of such size always contains \cite{deng2018arcface,de2018heterogeneous}.
More recently, the WebFace260M dataset was released \cite{zhu2021webface260m}.
This dataset is currently the largest public face recognition dataset.
It is composed of noisy 260M faces of 4M identities and a cleaned version composed of 42M faces and 2M identities.

\subsubsection{Evaluation Datasets}
While there is a plethora of old and small-scale face datasets for evaluation, we here focus only on the ones that suit our purpose best.
More datasets can, for example, be found online.\footnote{\url{http://face-rec.org/databases}}

With about 3312 images taken of 76 males and 60 females, AR face \cite{arface} is a relatively small dataset, but it is still in use today due to its unique face variations.
The images vary in facial expressions, illumination, and occlusion in the form of scarves and sunglasses.
Multi-PIE \cite{multipie} contains about 755'370 images shot in four sessions from 337 different subjects, covering 15 different camera view points, 19 different lighting conditions, and 7 distinct facial expressions.
SCface \cite{scface} contains 4160 images from 130 subjects taken by five video surveillance cameras of different qualities that were installed slightly above the head position.
Pictures of the participants were taken from three different distances where the smallest faces were just about 20 pixels in height.
The MOBIO dataset \cite{mobio} includes facial videos, images and speech recordings of 152 people taken with mobile devices over 12 different recordings.
This dataset is of particular interest since the view-point and background seen in the recordings are different from the default forward-facing images, and it provides two gender-dependent evaluation protocols.

Besides these datasets that allow to investigate different aspects of face recognition (occlusion, expression, and pose), other larger datasets are often used for evaluation.
The Good, the Bad \& the Ugly (GBU) dataset \cite{gbu} consists of 8'638 frontal images from 782 different identities.
It provides three protocols that mainly evaluate different illumination conditions called \textsf{Good}, \textsf{Bad}, and \textsf{Ugly}, where \textsf{Ugly} is the most difficult protocol, while \textsf{Good} is the easiest.
The IARPA Janus Benchmark C (IJB-C) \cite{ijbc} is currently the most widely used benchmark for face recognition.
IJB-C has the highest diversity in occlusion, occupation, and geographic origin, and image quality to better represent as much of the world's population as possible.
The dataset consists of a total of 31'334 images and 11'779 videos of 3'531 identities.

\subsection{Algorithms}

Before deep learning, face recognition was accomplished through traditional face recognition algorithms.
In \cite{gunther2016,gunther2017}, we surveyed and evaluated the performance of several traditional face recognition methods, where we used the open-source software Bob \cite{bob2012} and also published our code.\footref{fn:FRICE}
We showed that most of the traditional algorithms worked relatively well in good conditions but failed strongly when differences in facial expressions, illuminations, occlusions, and poses were evaluated.
The algorithm with the highest stability against most of these factors was found to be Inter-Session Variability (ISV) modeling \cite{wallace2011intersession}.

In recent years, deep learning has dominated and revolutionized the field of face recognition so that current face recognition surveys and reviews are full of deep learning methods.
These algorithms have advanced face recognition to a level that traditional methods can no longer reach \cite{mei2020}.
There are two main research directions in the academic community that have tried to improve the performance of neural networks, especially in unconstrained face recognition environments: the engineering of new network topologies and the definition of new loss functions.
Early versions of deep face recognition systems were using only a few convolutional layers, e.g., the VGGFace network \cite{vggface} used 13 such layers.
One of the most relevant contributions in terms of network topologies is the Residual Network \cite{he2015resnet}, which introduced residual connections between layers that allow training much deeper network structures than it was able before, the most common topologies have 18 to 152 layers.
Another architecture is the Squeeze and Excitation network \cite{squeezeandexcitation2017}, which integrates a special block into current network architectures that allows for automatic weighting of individual convolution channels.
Additionally, so-called lightweight network architectures have recently been developed.
One of these is the MobileNet \cite{howard2017mobilenets}, which uses depth-wise separable convolutions and neural architectural search to lead to a considerable reduction of parameters and, therewith, reduced the computing requirements compared to other networks with similar depth.

\begin{figure}[tb]
    \centering
    \subfloat[Original frontal \label{fig:original-notpreprocessed}]{\includegraphics[height=.09\textwidth]{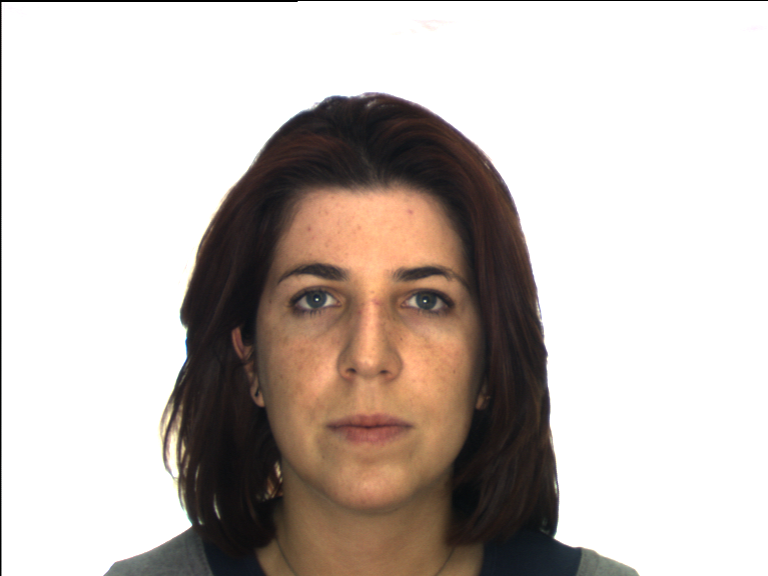}}
    \hspace*{0.03\textwidth}
    \subfloat[ArcFace frontal \label{fig:arcface-preprocessed}]{\includegraphics[width=.09\textwidth]{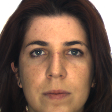}}
    \hspace*{0.03\textwidth}
    \subfloat[Facenet frontal \label{fig:vggfac2-preprocessed}]{\includegraphics[width=.09\textwidth]{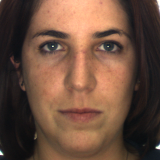}}
    \hfill
    \subfloat[Original profile \label{fig:original-notpreprocessed-profile}]{\includegraphics[height=.09\textwidth]{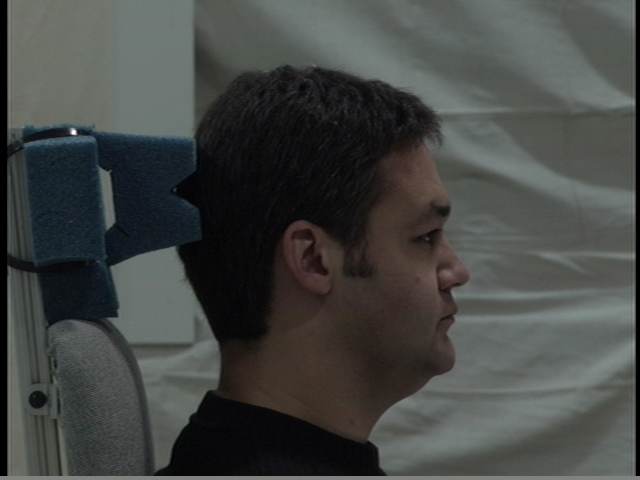}}
    \hspace*{0.03\textwidth}
    \subfloat[ArcFace profile \label{fig:arcface-preprocessed-profile}]{\includegraphics[width=.09\textwidth]{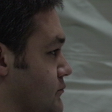}}
    \hspace*{0.03\textwidth}
    \subfloat[Facenet profile \label{fig:vggfac2-preprocessed-profile}]{\includegraphics[width=.09\textwidth]{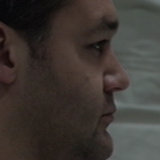}}
    \Caption[fig:preprocessing-examples]{Preprocessing Examples}{This figure shows for each network some preprocessed example images from the AR face and Multi-PIE datasets.
    }

\end{figure}

The most common loss function employed in classification tasks is the categorical cross-entropy loss used in combination with softmax activation, which is often called SoftMax loss.
The basic hypothesis of this loss is that the final embeddings (aka. deep features) that result from this closed-set end-to-end training are sufficiently discriminative for open-set problems, i.e., when subjects from the test datasets differ from the people used during training.
Several extensions on top of this basic hypothesis were created over the years, including Center loss \cite{centerloss}, which works in conjunction with the SoftMax loss to minimize the within-class distance of embeddings by learning a center for each class and penalizing the distance of features to the corresponding center \cite{hsu2020lossfunctions}.

\begin{figure*}[t]
    \centering
    \subfloat[neutral, occlusion, illumination \label{fig:arface-images}]{\includegraphics[width=.23\textwidth]{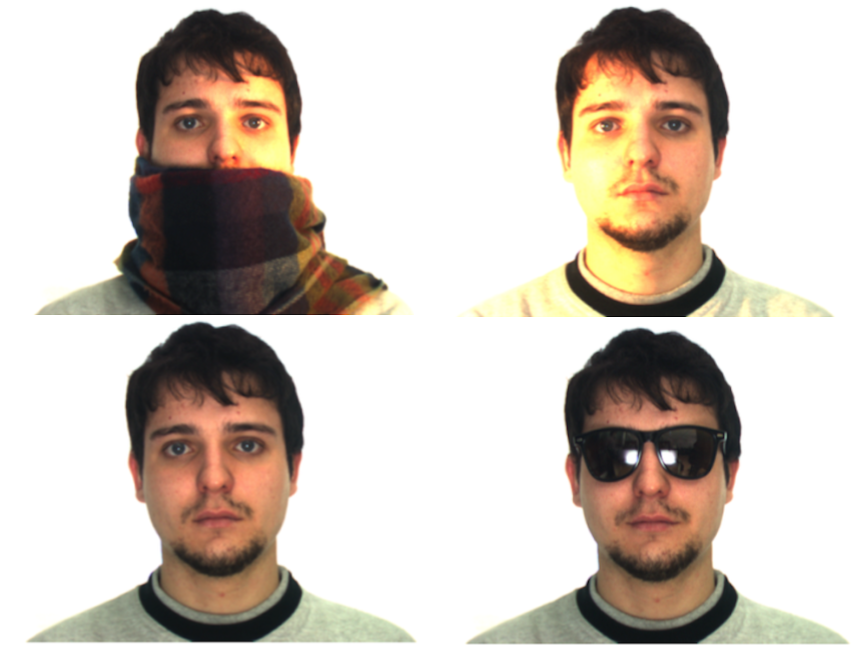}}
    \subfloat[Effect of illumination and occlusion \label{fig:occlusion-illumination}]{\includegraphics[page=2,width=.37\textwidth]{plot/arface_paper.pdf}}
    \subfloat[Effect of different occlusion types \label{fig:scarf-sunglasses}]{\includegraphics[page=1,width=.37\textwidth]{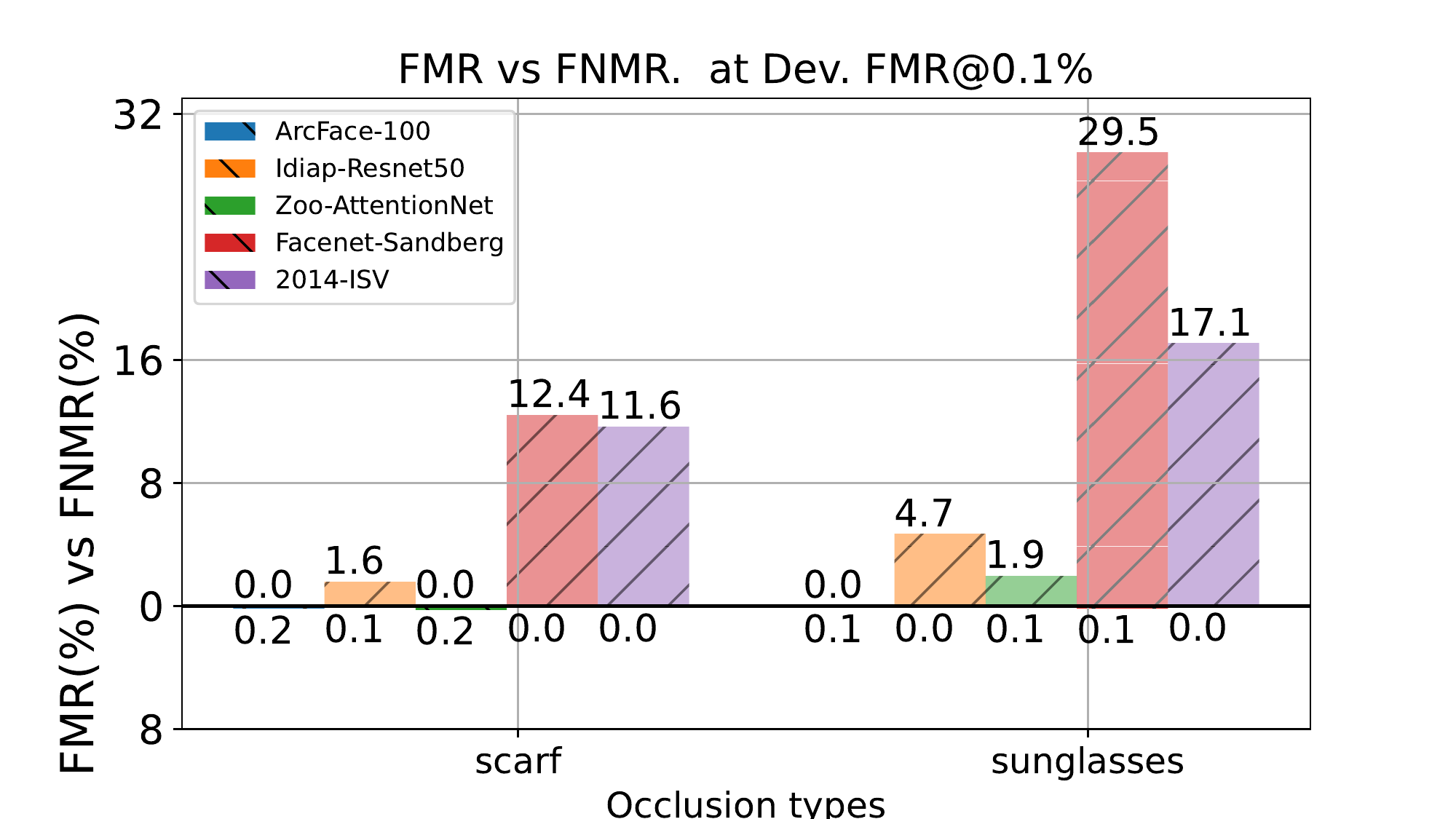}}
    \Caption[fig:occ]{Partial Occlusion}{
        Image examples of the AR face dataset and the effect of partial occlusion of the face on the tested algorithms.
        False Match Rates (FMR\%) and False Non-Match Rates (FNMR\%) on the \emph{eval} set are computed based on the threshold at FMR=0.1\% on the \emph{dev} set.
    }
\end{figure*}

Large-margin loss \cite{largemarginsoftmax}, also known as L-SoftMax, was one of the first loss functions that extended SoftMax with an angular margin, which was finally employed for face recognition by SphereFace \cite{sphereface}.
Later, CosFace \cite{cosface} extended this loss to use cosine similarity instead of angular losses and, finally, ArcFace \cite{deng2018arcface} introduced an additive angular margin to both maximize intra-class similarity and inter-class diversity.
The big advantage of this margin is that it allows some similarity between faces of different people and does not force all of them to be as dissimilar as possible.
Latest loss functions include AdaCos \cite{zhang2019adacos}, PS2Grad \cite{zhang2019p2sgrad}, ring loss \cite{zheng2018ring}, and MagFace \cite{meng2021magface}.
Also, loss functions that explicitly tackle the issue of open-set face recognition have been proposed \cite{guenther2020watchlist}, but those techniques will not be discussed further in our study.
Other losses that are worth mentioning are Triplet \cite{facenet2015} and Contrastive Loss \cite{chopra2005learning}, which are metric learning approaches that work directly on the embedding space by explicitly minimizing within-class and maximizing between-class variability.

%% file: approach.tex
\section{Evaluation procedure}

The experiments described in the present work all rely on the software Bob \cite{bob2012,bob2017}, an open-source signal processing and machine learning toolbox.
Particularly, we make extensive use of its Biometric Recognition Pipelines \cite{guenther2012facereclib}, which are \emph{easily extensible to use new face recognition algorithms based on deep learning} and allow an easy way of reproducing\footnote{We are aware that some of the employed datasets are no longer publicly available -- and we are not allowed to share the data ourselves -- which limits the reproducibility of some of our experiments. We provide the resulting recognition scores for further evaluation.} experiments \cite{bob2017}.
In our current evaluation we will make use of the newly added interfaces for running experiments with deep networks \cite{linghu_zhang2021master_project} and the work of \cite{schmidli2021bachelor}.

\fig{FR-process} illustrates the three different steps in the biometric recognition process of Bob.\footnote{\url{https://www.idiap.ch/software/bob/docs/bob/bob.bio.base/v5.0.0/biometrics_intro.html}}
The biometric \textbf{Dataset} stores all information required to run a biometric recognition process, which are the original images and their identity labels, facial landmarks used for alignment, and the evaluation protocol that defines which images should be compared.
The \textbf{Transformer} is essentially a scikit-learn Pipeline\footnote{\url{https://scikit-learn.org/stable/modules/generated/sklearn.pipeline.Pipeline.html}} containing a sequence of steps to process a sample.
Such a pipeline can assume a different sequence of steps depending on the biometric algorithm.
In the case of face recognition it is usually composed of a face and facial landmark detector, face alignment, and a feature extractor.
Since in this work we are evaluating face recognition algorithms and not facial landmark detectors, we replace the face detector by using hand-annotated landmarks for the alignment step in our experiments.
Finally, the \textbf{Biometric Algorithm} has functions to enroll a client (create a biometric template) and compute a similarity score between a given template and probe sample.
When several images are used for enrollment, the simple average of the embeddings is computed.

\subsection{Evaluated Algorithms}
Bob's face recognition package\footnote{\url{http://gitlab.idiap.ch/bob/bob.bio.face}} has more than 30 different face recognition systems available (including traditional methods based on hand-crafted features, as well as many modern deep-learning algorithms and pre-trained networks) ready to be used.
Because of page limits, in this work we will make use of four different deep-learning-based face recognition systems available in Bob.
However, scores from all the available systems will be available.\footref{fn:package}
The four systems are the following -- in chronological order of publication:
The first system is the \textbf{Facenet-Sandberg}\footnote{\url{https://github.com/davidsandberg/facenet}} trained using a pruned version of the MS-Celeb-1M dataset using the Inception-ResNet-v1 backbone \cite{facenet2015}.
The second network is the ArcFace model \cite{deng2018arcface} from InsightFace (\textbf{ArcFace-100}) that is based on ResNet-101 architecture and trained using the ArcFace loss.
The third network is taken from the FaceX-Zoo models \cite{wang2021facex}, which contain several face recognition systems which are all integrated in Bob; in this work we have used the \textbf{Zoo-AttentionNet} backbone.
The forth method is based on ResNet-50 architecture trained using the ArcFace loss (\textbf{Idiap-Resnet50}) on a pruned version of the MSCeleb-1M dataset.
Finally, to show the improvement made over the last eight years, we also provide results of the top-performing method of our old evaluation, i.e., the Inter-Session Variability (ISV) modeling of Discrete Cosine Transform (DCT) features \cite{wallace2011intersession}.

\begin{figure*}[t!]
    \centering
    \subfloat[Facial expressions examples: neutral, smile, surprise, squint, disgust, scream\label{fig:multipie-expression-img}]{\includegraphics[width=1.0\textwidth]{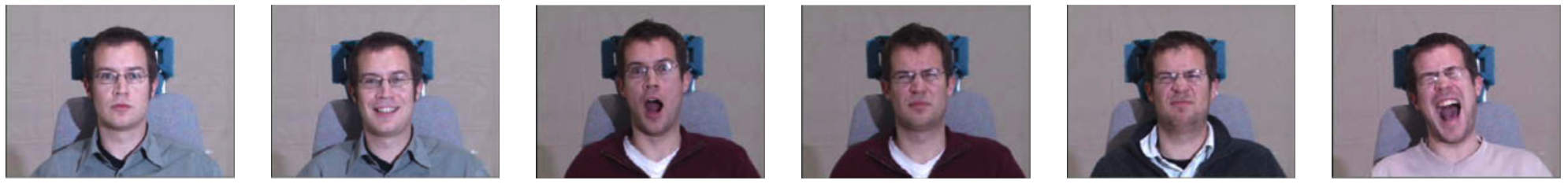}}

    \subfloat[Poses examples from -90 to 90 in steps of 15 degrees\label{fig:multipie-pose-img}]{\includegraphics[width=1.\textwidth]{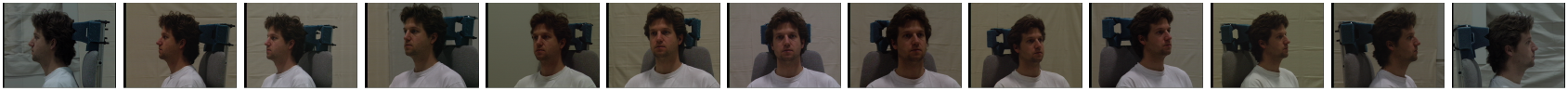}}

    \subfloat[Effect of different expressions \label{fig:expression}]{\includegraphics[page=1,width=.50\textwidth]{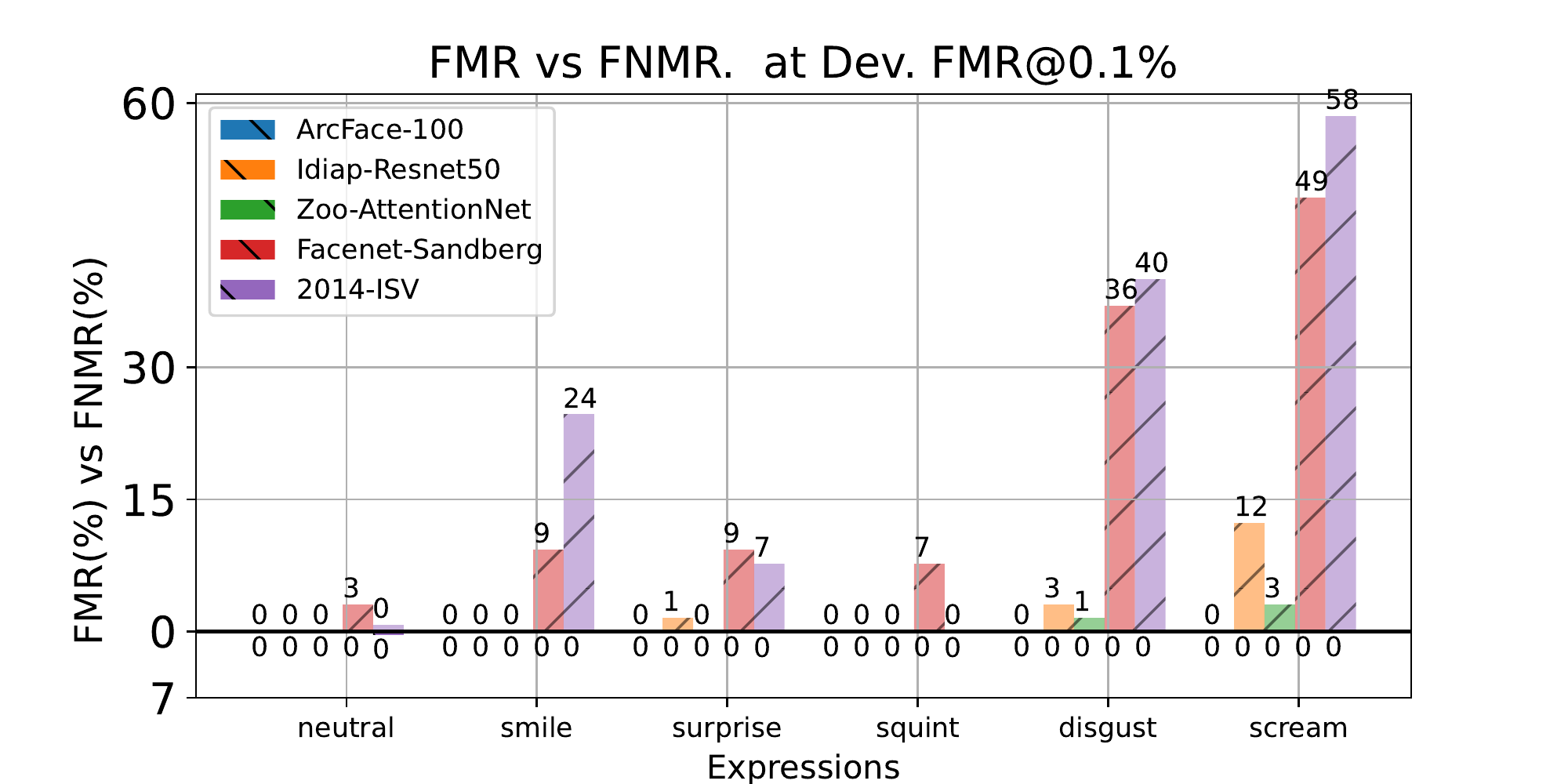}}
    \hspace*{0.05\textwidth}
    \subfloat[Effect of different poses \label{fig:poses}]{\includegraphics[page=1,width=.50\textwidth]{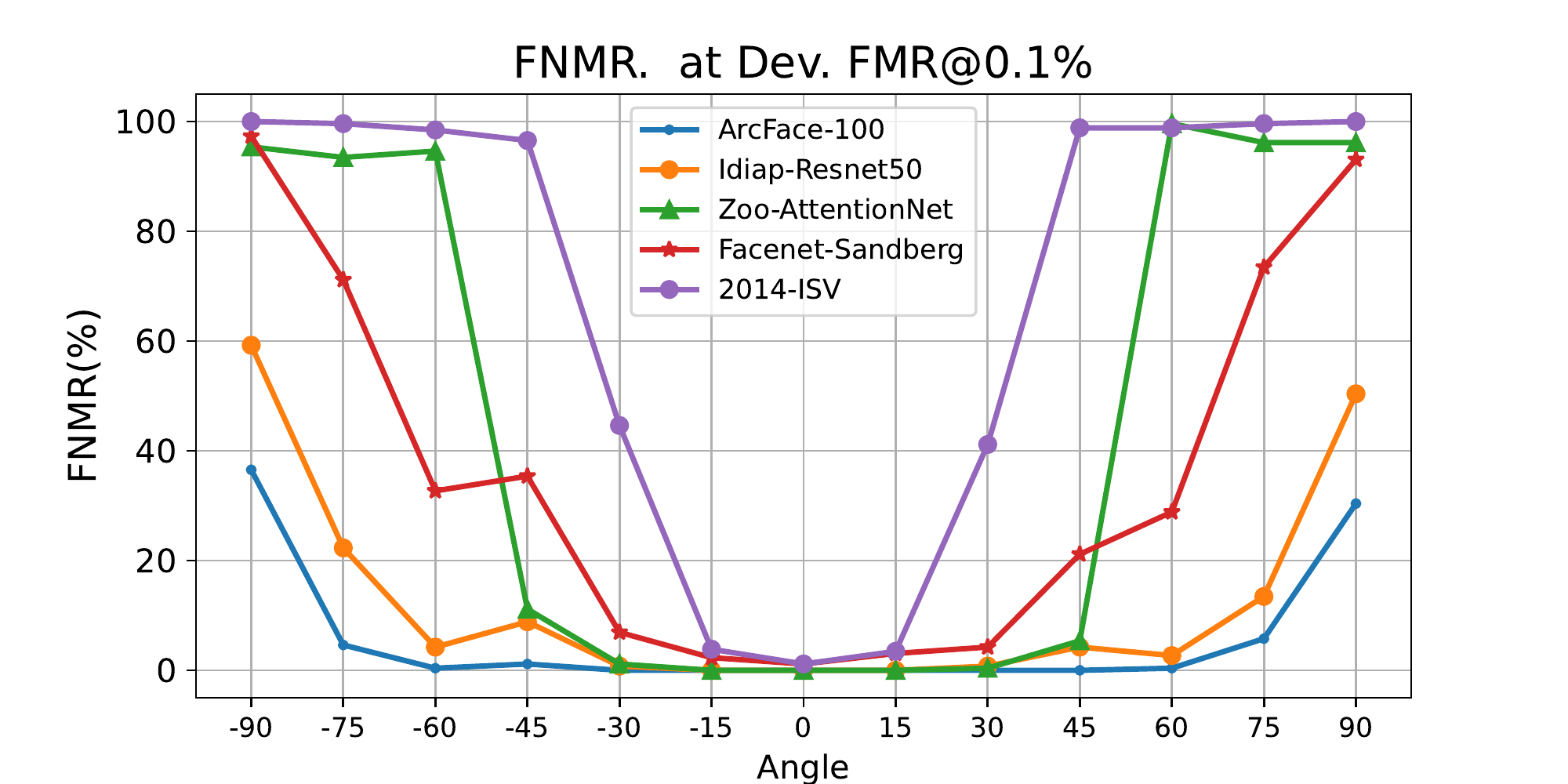}}
    \Caption[fig:exp-pose]{Expression and Pose}{This figure shows the effect of different expressions and poses of the face on the tested neural networks. Example images for facial expressions and poses are displayed in subfigure \protect\subref*{fig:multipie-expression-img} and subfigure \protect\subref*{fig:multipie-pose-img}.}
\end{figure*}

\subsection{Image Preprocessing}
Preprocessing plays a crucial role in the face recognition process as it can affect the performance of feature extraction networks \cite{wartmann2021bachelor}.
For all datasets used in this work -- except for IJB-C -- hand-labeled facial landmarks are available, which can be used to align faces directly according to the desired size and the location of these landmarks in the target images.
Unfortunately, many research papers lack detailed information on how preprocessing is done, making it particularly difficult for others to reproduce experiments.
For ArcFace, only the required input image dimension of $112\times112$ is provided \cite{deng2018arcface}.
There were also some scripts to align faces based on landmarks detected with a MTCNN \cite{mtcnn}, but our hand-labeled landmarks do not correspond to those extracted by MTCNN and, thus, it was not entirely clear how to achieve alignment.
Six pictures, which indicate some kind of alignment, could be taken from their GitHub\footnote{\url{https://github.com/deepinsight/insightface}} repository, which we used to manually estimate average landmark locations in the target images.
In a similar manner, face images are cropped to $160\times160$ for Facenet.

The alignment for almost all experiments in this work has been done based on eye landmarks.\footnote{See our \href{https://www.idiap.ch/software/bob/docs/bob/docs/stable/bob/bob.ip.base/doc/guide.html\#normalizing-face-images-according-to-eye-positions}{face alignment guide}}
The experiments on the Multi-PIE dataset required an additional alignment point since the images do not always provide two visible eyes \cite{multipie}. In these cases, the visible eye and the respective corner of the mouth served as a reference point.
Defining these landmarks was even more cumbersome and required trail-and-error executions of algorithms \cite{wartmann2021bachelor}.
While our experiments indicate that these landmarks work approximately well, there is no guarantee that these are the best landmark locations to be used.
Examples of preprocessed images can be found in \fig{preprocessing-examples}, while exact landmark locations in the aligned images are given in \tab{preprocessing}.

\label{sec:ImagePreprocessing}

\subsection{Evaluation Criteria}
With the computed similarity scores from our face recognition pipeline, several evaluations are possible.
To evaluate verification protocols, usually the False Match Rate (FMR) and the False Non-Match Rate (FNMR) are computed based on a certain similarity score threshold.
By varying this threshold, Receiver Operating Characteristics (ROC) can be plotted and compared.
Unfortunately, however, ROCs have the issue that they are computed over the test set and any threshold selected on an unseen set of images is not guaranteed to perform equally well in practice.

For this reason, unless the datasets provide different standard evaluation procedures, we have split all of our evaluation protocols into two groups of non-overlapping identities.
The first group \emph{dev} is used to compute a threshold based on some criteria, and this threshold is then applied to the second group \emph{eval} to compute the final performance metrics.
In our previous study \cite{gunther2016,gunther2017}, we mainly computed the threshold based on the Equal Error Rate (EER) on \emph{dev} and reported Half Total Error Rates (HTER) on \emph{eval}.
Since most security-relevant applications of face recognition want to assure a very small risk of imposters being recognized as genuine, more reasonable thresholds are rather computed for low FMR values.
Therefore, in our current evaluation we have switched to compute the threshold based on an FMR of 0.001 (or 0.1\,\%) on the \emph{dev} set.
Notably, when running an evaluation with several sub-protocols (\sec{results}) we compute a \textbf{single} threshold for the \textbf{combined} scores over all protocols on the \emph{dev} set.
Finally, we report both FMR and FNRM on the \emph{eval} set for each sub-protocol separately.

Open-set identification systems are evaluated using the Detection and Identification Rate Curve \cite{jain2011handbook}, which is also called the Open-Set ROC and is the standard metric in NIST evaluations.\footnote{\url{https://www.nist.gov/programs-projects/face-recognition-vendor-test-frvt-ongoing}}
For consistency, we will use the NIST terms and evaluate the False Positive Identification Rate (FPIR) and True Positive Identification Rate (TPIR) at rank 1 based on a certain similarity threshold.
By varying this threshold, the TPIR can be plotted over the FPIR.
Note that the closed-set identification performance can be obtained at FPIR=1, i.e., on the right-hand side of the plot.
While we are aware that this measure has the same deficits as the verification ROC discussed above, we leave the development of better-suited open-set evaluation metrics for future work.

%% file: experiments.tex
\section{Experiments}
\label{sec:results}

In the following, we present the results of all face recognition experiments performed using the four networks \textbf{Facenet-Sandberg}, \textbf{ArcFace-100}, \textbf{Zoo-EfficientNet}, and \textbf{Idiap-Resnet50}.
Furthermore, we have included experiments from the best overall baseline \cite{wallace2011intersession} from our previous publications \cite{gunther2016,gunther2017}, which we have marked as \textbf{2014-ISV}.
The scores from other systems evaluated in our old study are also available for comparison.\footnote{\label{fn:resources}See the FRICE 2016 section in \url{https://www.idiap.ch/webarchives/sites/www.idiap.ch/resource/biometric}}
In this way, we have a perspective on which fronts current state-of-the-art systems improved.

\subsection{Face Variations}
\label{sec:variations}
First, the algorithms are tested against three types of face variations, more precisely partial occlusion, different expressions and poses.
In all evaluated datasets in this section, gallery templates are enrolled from neutral faces, i.e., images with neutral frontal illumination showing a face in frontal pose and with neutral facial expression.
On the other hand, probe images are equipped with one of the above-mentioned variations.
This assures that recognition does not happen because gallery and probe share the same variation, but only because the algorithm is able to ignore the variation and still can recognize the person.

\subsubsection{Partial Occlusion}

Partial occlusion is a common issue in unconstrained face recognition environments, which makes the recognition of identities harder.
Especially during the COVID-19 pandemic, when this work was written, many people wore masks that covered their faces from chin to nose, which has been shown to influence face recognition \cite{damer2020effect}.
The AR face dataset \cite{arface} is used to evaluate the performance of the algorithms with respect to different partial occlusions.
It consists of four protocols \textsf{expression}, \textsf{occlusion}, \textsf{illumination}, and \textsf{occlusion\_and\_illumination}, \sfig{arface-images} displays some example images from the used protocols.
For all experiments on this dataset, only images with neutral facial expressions were used to observe the influence of occlusion and illumination as isolated as possible.
The identities were split up into 24 males and 19 females for each \emph{dev} and \emph{eval} sets.

\begin{figure}[tb]
    \centering
    \subfloat[Enrollment\label{fig:scface:enroll}]{\ig[.08]{SCface/001_frontal}}\,
    \subfloat[Close\label{fig:scface:close}]{\ig[.08]{SCface/001_cam1_3}}\,
    \subfloat[Medium\label{fig:scface:medium}]{\ig[.08]{SCface/001_cam1_2}}\,
    \subfloat[Far\label{fig:scface:far}]{\ig[.08]{SCface/001_cam1_1}}

    \subfloat[Results\label{fig:scface:res}]{\igp[.45]{1,trim=0 0 0 20,clip}{scface_dnn_new}}
    \Caption[fig:scface]{SCface}{This figure displays some example images of the SCface dataset and the performance of the tested neural networks.
    }
\end{figure}

In \sfig{occlusion-illumination} we present the False Non-Match Rate (FNMR) and False Match Rate (FMR) on the \emph{eval} set using the score threshold at FMR at 0.1\% in the \emph{dev} set.
As can be seen, most of the networks are not severely affected by occlusion or illumination.
The more recent \textbf{ArcFace-100}, \textbf{Zoo-AttentionNet}, and \textbf{Idiap-Resnet50} present both FNMR around 1\% and FMR around the operational threshold of 0.1\%.
The slightly older \textbf{Facenet-Sandberg} presents some difficulties with different types of occlusion that are isolated in \sfig{scarf-sunglasses}, e.g., an FNMR of 29\% is reported for sunglasses.
Our selected top three algorithms present significant improvements in terms of FMR and FNMR compared with our \textbf{2014-ISV} system from eight years ago, which is severely impacted by both occlusion and illumination.

\begin{figure*}[t]
    \centering
    \subfloat[Good \label{fig:gbugood}]{\includegraphics[page=1,width=.31\textwidth]{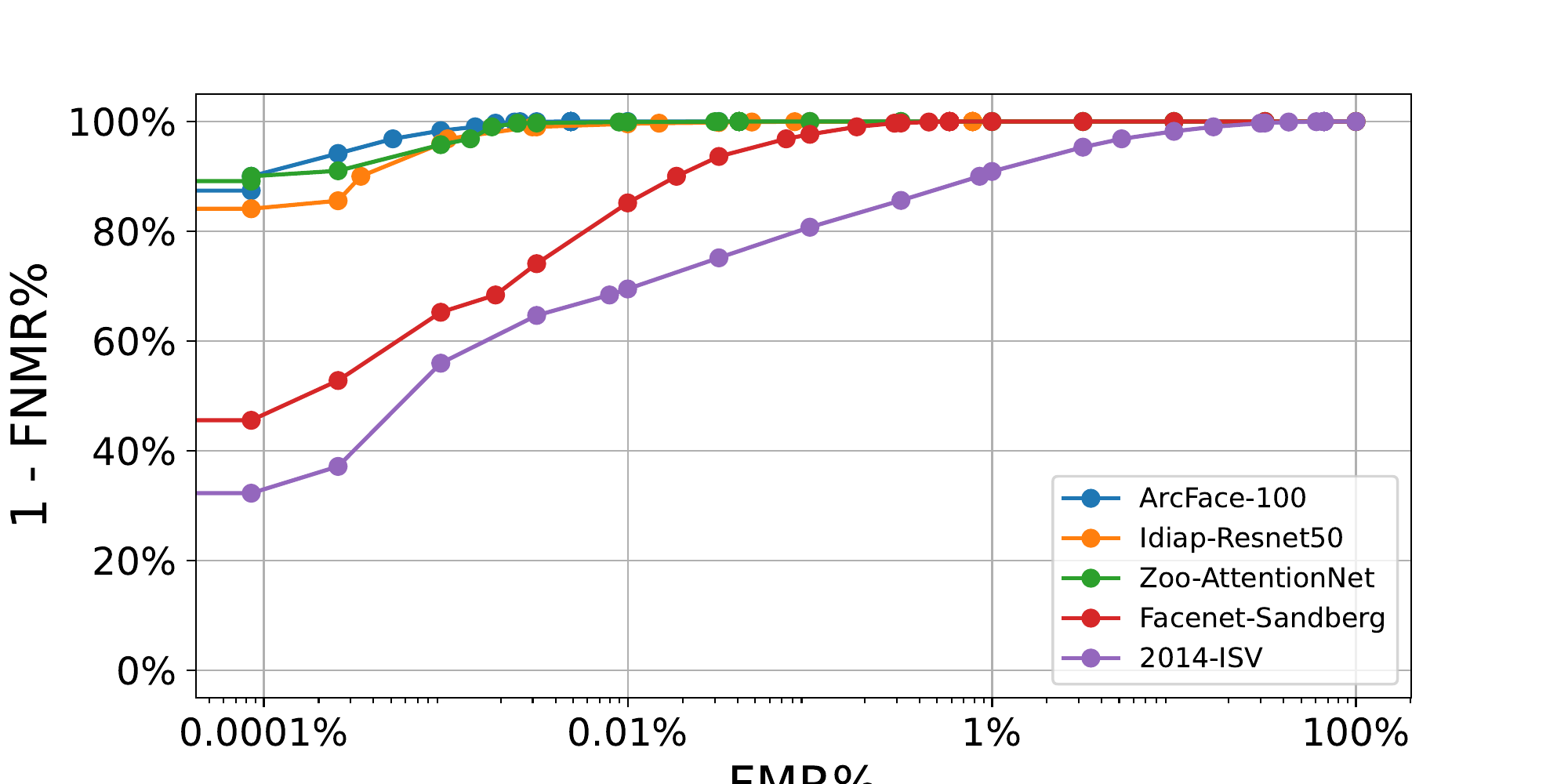}}
    \hspace*{0.03\textwidth}
    \subfloat[Bad \label{fig:gbubad}]{\includegraphics[page=1,width=.31\textwidth]{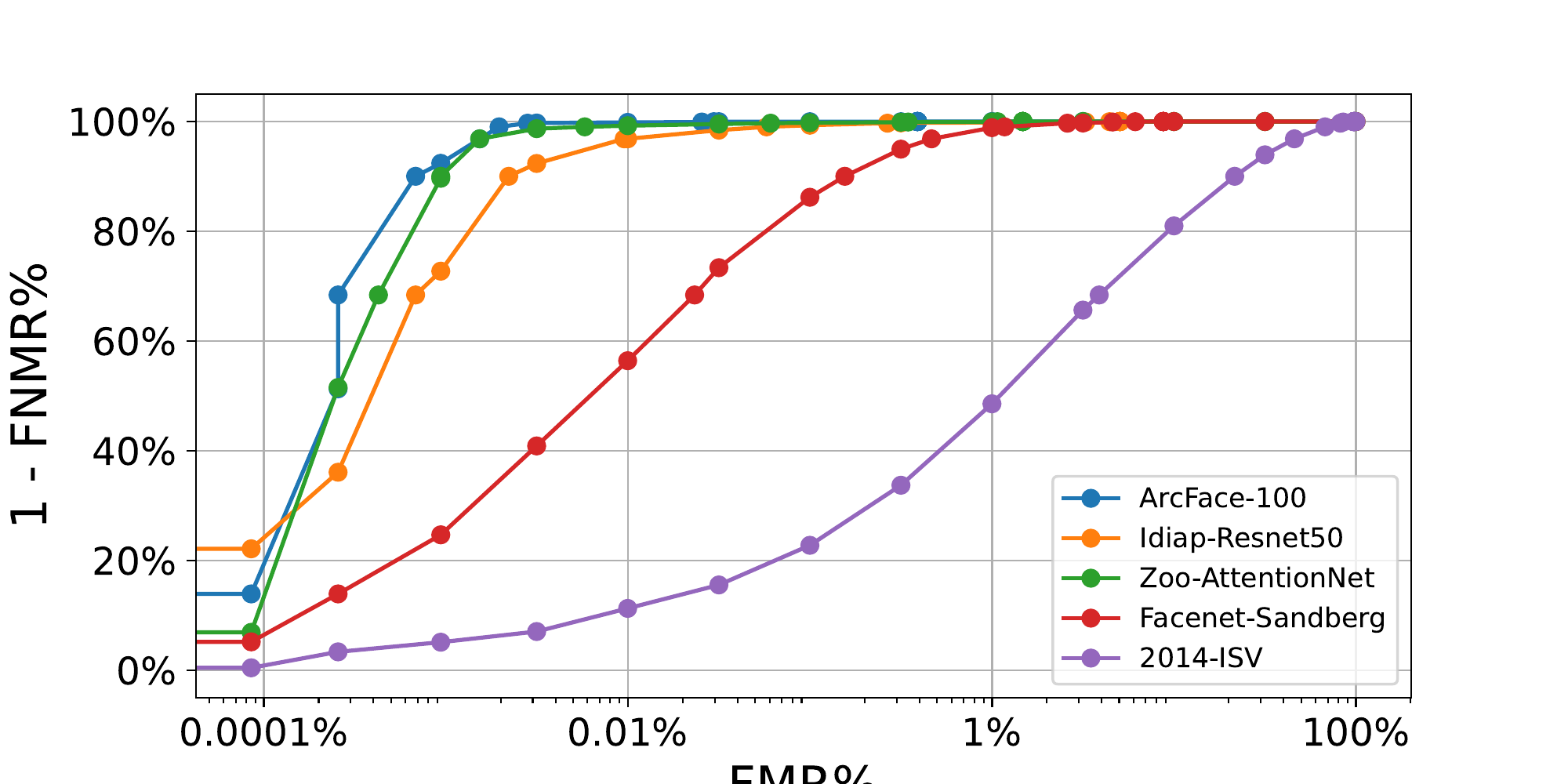}}
    \hspace*{0.03\textwidth}
    \subfloat[Ugly \label{fig:gbuugly}]{\includegraphics[page=1,width=.31\textwidth]{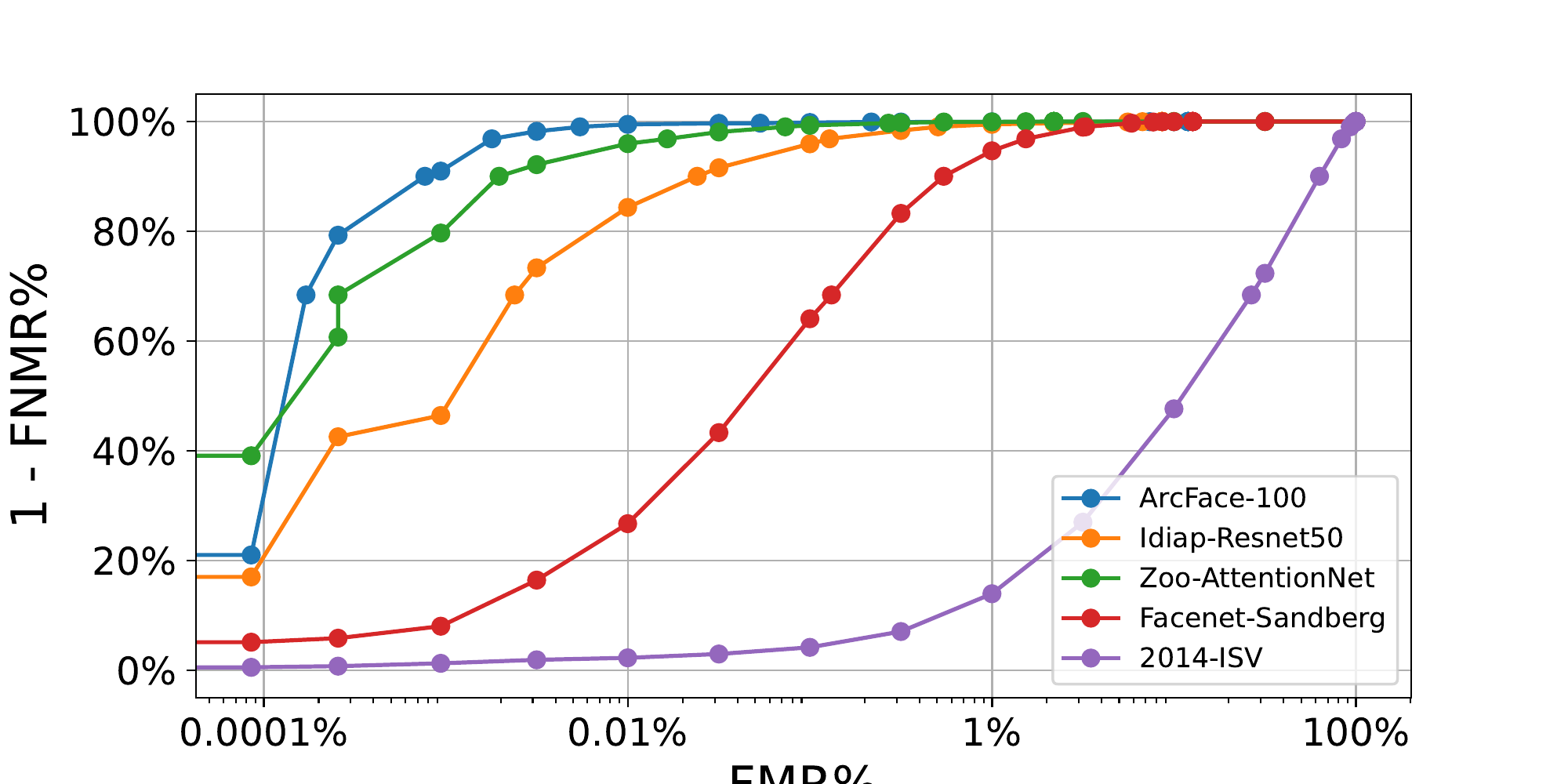}}
    \Caption[fig:gbu]{The Good, the Bad \& the Ugly}{ROC curves for the protocols \textsf{Good}, \textsf{Bad} and \textsf{Ugly} of the GBU dataset.}
\end{figure*}

\subsubsection{Facial Expressions}
Humans are emotional beings and tend to show their emotions intensely through facial expressions, which has a significant visual impact on facial features \cite{guo2019survey}.
Therefore, modern face recognition algorithms must be able to handle a wide range of facial expressions.
The Multi-PIE \cite{multipie} dataset with its protocol \textsf{E} is used to test the algorithms against a variety of expressions seen in \sfig{multipie-expression-img}.
64 identities are used in the \emph{dev} set, and the \emph{eval} set is composed of 65 identities.
Five faces per identity with neutral expressions were considered for gallery template enrollment.

\sfig{expression} shows the FNMR and FMR for different expressions by setting the decision threshold at FMR at 0.1\% in the development set.
The plot reveals that most recent networks can handle facial expressions well.
All systems present an FMR around the operation threshold, which is an indication of homogeneity of both development set and evaluation sets.
\textbf{ArcFace-100} and \textbf{Zoo-AttentionNet} show FNMR around 1\% for all expressions.
The \textbf{Idiap-Resnet-50} presents an FNRM of around 1\% for all expressions except for ``scream'' whose FNMR drastically increases to 12.3\%.
Both \textbf{Facenet-Sandberg} and \textbf{2014-ISV} present high FNRM for almost all expressions, reaching the highest values of 49\% for and 59\% for the ``scream'' expression.


\subsubsection{Face Poses}
Another aspect that challenges face recognition is the presence of different face poses.
It is known that the performance of neural networks significantly drops when faces are no longer frontal \cite{sengupta2016}.
Protocol \textsf{P} from the Multi-PIE dataset is used to observe the performance of neural networks on pose variations.
This protocol provides faces rotated from left to right in steps of 15 degrees of yaw angle, examples can be seen in \sfig{multipie-pose-img}.
The facial expressions are neutral, without any type of occlusion or strong illumination.
When both eyes are visible, i.e. for $\pm45\degree$, the hand-labeled eye positions are used for alignment.
For the poses with deviations of more than $45\degree$ from frontal, only one eye is visible and, therefore, the alignment used the visible eye and the corner of the mouth.
64 identities are used for the \emph{dev} set and 65 for the \emph{eval} set, and five frontal images per identity are enrolled in a gallery template.

\sfig{poses} shows the FNMR for different pose angles.
For this experiment, we observed a similar trend in terms of FMR as in the previous one (all FMR are around the operational threshold).
For that reason, we are plotting only the values of FNMR.
It can be observed that all systems are able to handle well frontal poses with an angle of less than $\pm15\degree$.
FNMR starts to drastically increase for the \textbf{2014-ISV} for angles larger than $\pm30\degree$, while all modern systems present similar FNMR for this particular set of angles.
For $45\degree$ angles, the FNMR of \textbf{Facenet-Sandberg} starts to increase to around 35\,\%, while \textbf{Arcface-100} still presents an FNMR 0\,\% for this angle and \textbf{Idiap-Resnet50} and \textbf{Zoo-AttentionNet} increase to an FNMR of around 6\%.
For angles above $\pm45\degree$ the FNMR of \textbf{Zoo-AttentionNet} drastically increases to around 100\%, while the \textbf{Idiap-Resnet50} and \textbf{Arcface-100} slowly increase to around 60\% and 35\% for angles of $\pm90\degree$.

\subsubsection{Face Sizes}

The surveillance camera face (SCface) dataset \cite{scface} contains images taken by different low-resolution video surveillance cameras at three different distances.
The three protocols \textsf{close}, \textsf{medium}, and \textsf{far} are used to evaluate the performance on different camera distances.
For each protocol, images of 44 identities are used for the \emph{dev} set, and 43 for the \emph{eval} set.
One frontal image taken in passport quality as shown in \sfig{scface:enroll} is used for model enrollment, which differs dramatically in quality from the probe images, e.g., the \textsf{far} probe face as shown in \sfig{scface:far} has only about 20 pixels of height.

\sfig{scface:res} shows the FMR and FNMR on the \emph{eval} set, indicating FMR values close to the estimated operational threshold.
In terms of FNMR we could observe that for short distances the \textbf{ArcFace-100} is the best system, presenting 0\%, followed by \textbf{Idiap-Resnet50} and \textbf{Zoo-AttentionNet} with 1.4\% and 4.7\% respectively.
With around 20.9\%, \textbf{Facenet-Sandberg} presents a very high FNMR.
Once the distance between the probe subject and the camera increases, decreasing the image resolution, the FNRM also increases.
At long distances (\textsf{far}), the \textbf{ArcFace-100} and the \textbf{Zoo-AttentionNet} present an FNMR of $\approx65\%$.
For the \textbf{Facenet-Sandberg} and \textbf{Idiap-Resnet50} FNMR reaches above 90\%.
Low-resolution probe samples seem to have a substantial impact on the performance of the algorithms.

\subsection{Unconstrained Evaluations}
This section provides the results for some common datasets, some of which were also evaluated in \cite{gunther2017}, including MOBIO, GBU, and IJB-C.

\subsubsection{MOBIO}
The MOBIO dataset consists of video frames taken via mobile phone or laptop from \textsf{male} and \textsf{female} subjects.
The \emph{dev} set consists of 18 females and 24 males with 1'890 and 3'520 images, respectively, while the \emph{eval} set contains 20 females and 38 males with 2'100 and 2'990 images. 
Five images per person are used for model enrollment. 

\fig{Mobio} shows the FMR and FNMR on the \emph{eval} set.
\textbf{ArcFace-100}, \textbf{Idiap-Resnet50}, and \textbf{Zoo-AttentionNet} present an FNMR below 0.2\% for all genders.
The \textbf{Facenet-Sandberg} presents a slightly higher FNMR compared with the others, around 2\% for both genders.
These FNMRs present a substantial improvement compared with the \textbf{2014-ISV}, which reaches around 40\%, and for which also the FMR raises substantially.

\subsubsection{The Good, the Bad, \& the Ugly}

\begin{figure}[t]
    \centering
    \igp[.5]{1,trim=0 0 0 20,clip}{mobio_dnn_new}
    \Caption[fig:Mobio]{MOBIO}{This figure shows the performance of the tested neural networks on the MOBIO dataset.}
\end{figure}

The experiments on the GBU \cite{gbu} dataset are performed on all three protocols \textsf{Good}, \textsf{Bad}, and \textsf{Ugly}.
Since the default evaluation protocols are not split into \emph{dev} and \emph{eval} sets, we are only able to report the ROC curve.
The model enrollment uses only one image per identity, but there are several models per identity defined by the protocol.

All networks performed well on protocol \textsf{Good}, with the exception of \textbf{Facenet-Sandberg}.
The best performance was achieved with the \textbf{ArcFace-100}, which provides the best FNMR for all FMR operational points.
For the protocol \textsf{Bad}, \textbf{ArcFace-100} performed the best, while the \textbf{Idiap-Resnet50} had a slight decrease for FMR below 0.01\%.
The same trends could be followed by the protocols \textsf{Ugly}.
Interestingly, for very low FMR values, results on the \textsf{Ugly} protocol are even better than on the \textsf{Bad} protocol, which indicates that the definition of Bad and Ugly has shifted since the development of this dataset.

\begin{figure*}[t!]
    \centering
    \subfloat[Example of enrollment samples
    \label{fig:ijbc_enroll}]{\includegraphics[page=1,width=.24\textwidth]{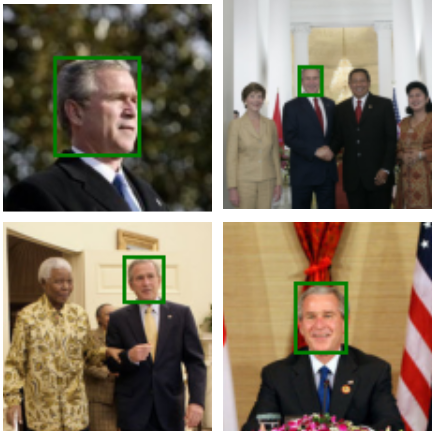}}
    \hspace*{0.03\textwidth}
    \subfloat[Example of probe samples \label{fig:ijbc_probe}]{\includegraphics[page=1,width=.24\textwidth]{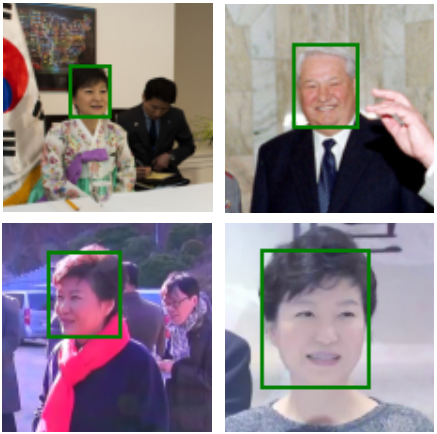}}
    \hspace*{0.03\textwidth}
    \subfloat[Open-Set ROC Curve \label{fig:ijbc_dir}]{\includegraphics[page=1,width=.35\textwidth]{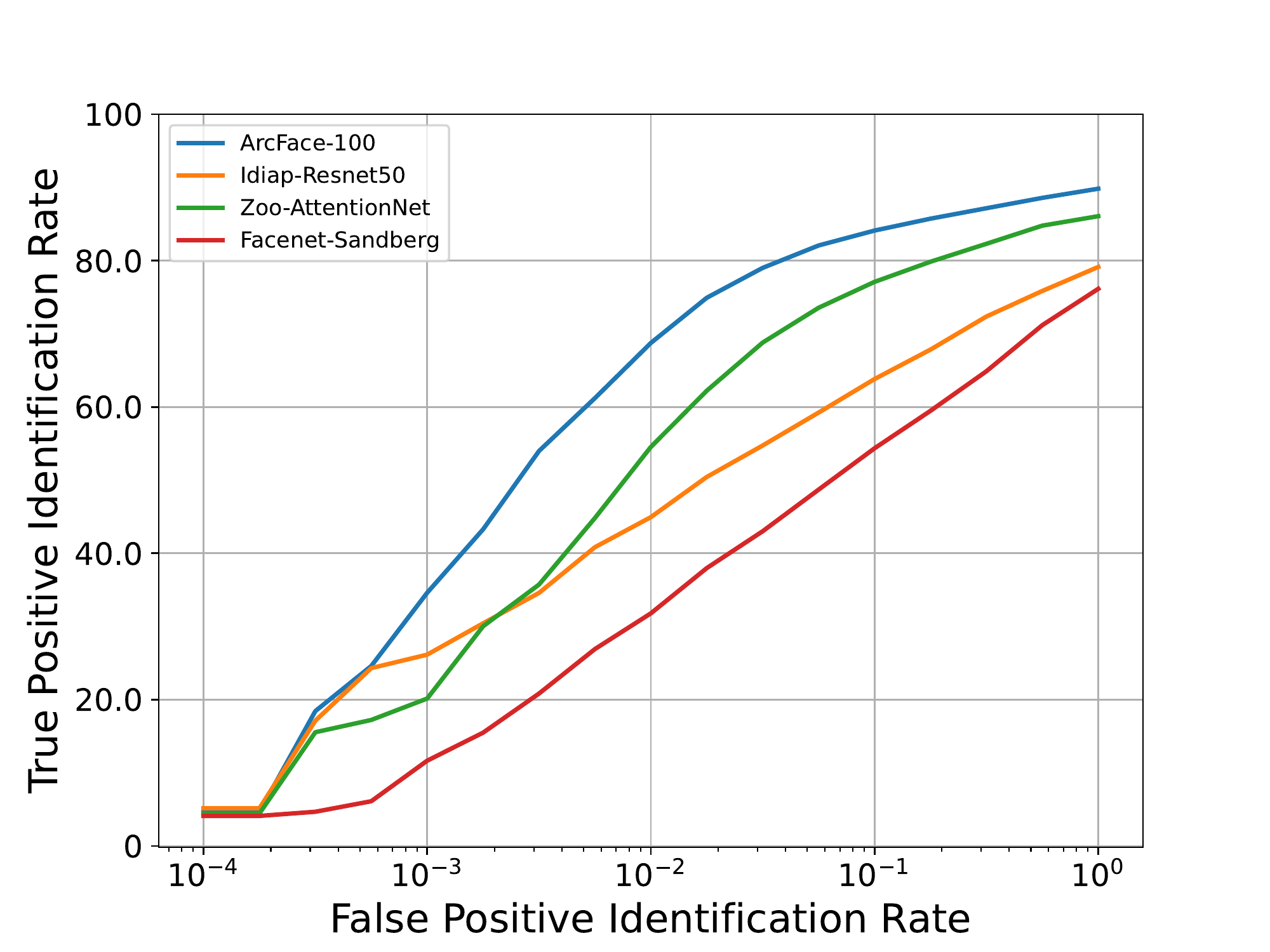}}
    \Caption[fig:ijbc]{IJB-C}{This figure displays some example images of the IJB-C dataset and the Open-Set ROC curve.}
\end{figure*}
\subsubsection{IJB-C}

The IARPA Janus Benchmark C (IJB-C) is one of the most challenging evaluation datasets in face recognition research.
This dataset contains evaluation protocols for face detection, face clustering, face verification, and open-set face identification.
In this work we focused on open-set evaluation and as such, we will use the protocol \textsf{test4-g1}.
This protocol contains a gallery of 1'170 subjects and a set of 1'759 ``unknown'' probes.
Since IJB-C does not provide eye locations for face alignment, we first cropped the faces according to a slightly enlarged ground-truth face bounding box, detected the facial landmarks via MTCNN, and used the detected eye locations for alignment according to \tab{preprocessing}.

It is possible to observe that in this setup, \textbf{ArcFace-100} presents the bests TPIR, followed by the \textbf{Zoo-AttentionNet}.
For closed-set results, i.e. TPIR=1 at the right-hand side of the plot, identification rates up to 90\,\% can be reached by the best-performing network.
However, the TPIR rapidly decreases with decreasing FPIR.
For FPIR=0.001 ($0.1\,\%$) all systems operate with a TPIR well-below 40\,\%.
This means, to have a level of false positives of one in a thousand, we should expect an identification rate of 40\,\%.
This is far from being practical in any surveillance camera application.

%% file: discussion.tex
\section{Discussion}

\begin{figure*}[t!]
    \centering
    \subfloat[Protocol-Specific Threshold on \emph{dev}\label{fig:thresholds:optimal}]{\includegraphics[width=.45\textwidth]{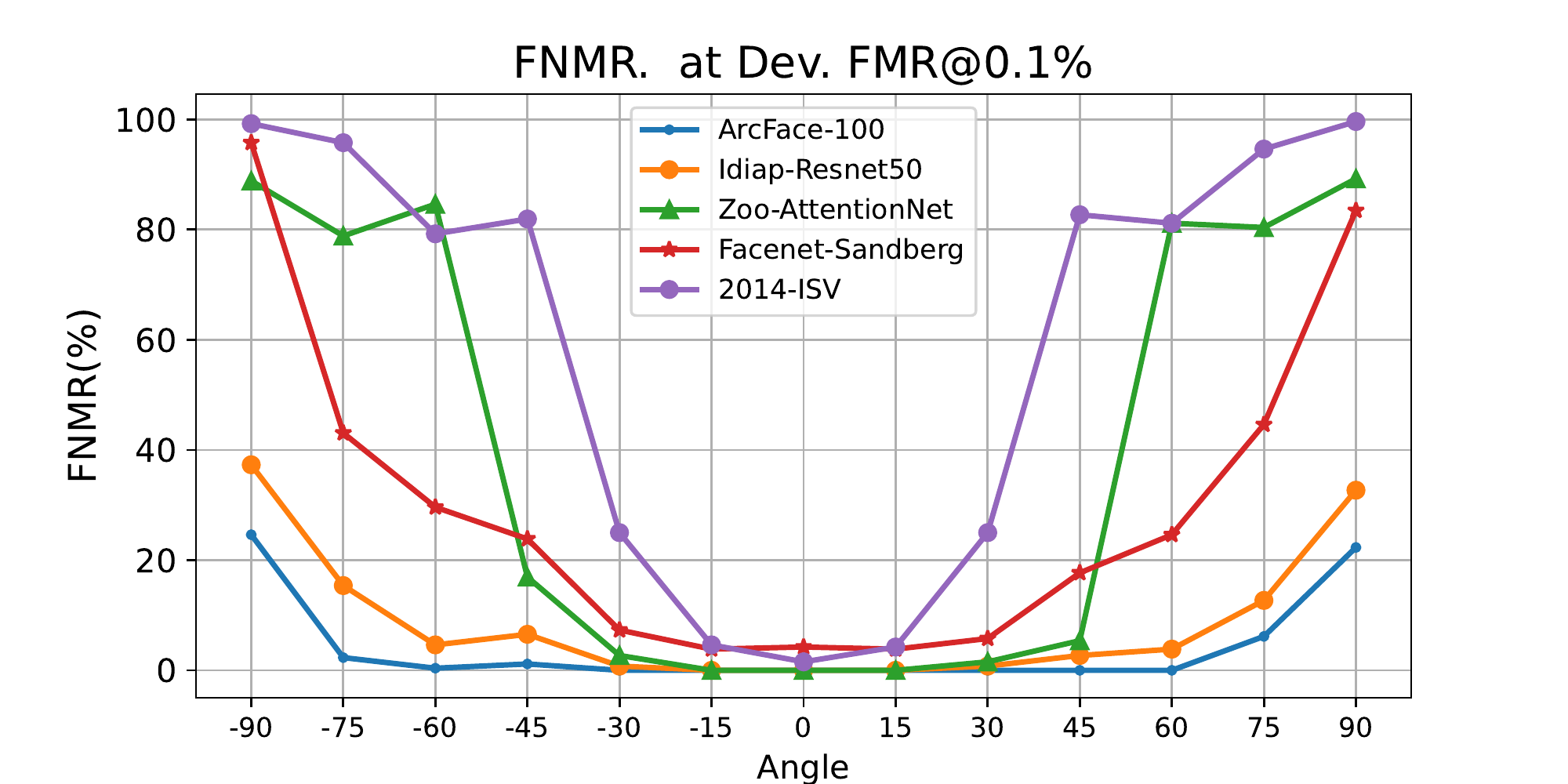}}
    \subfloat[Protocol-Specific Threshold on \emph{eval}\label{fig:thresholds:eval}]{\includegraphics[width=.45\textwidth]{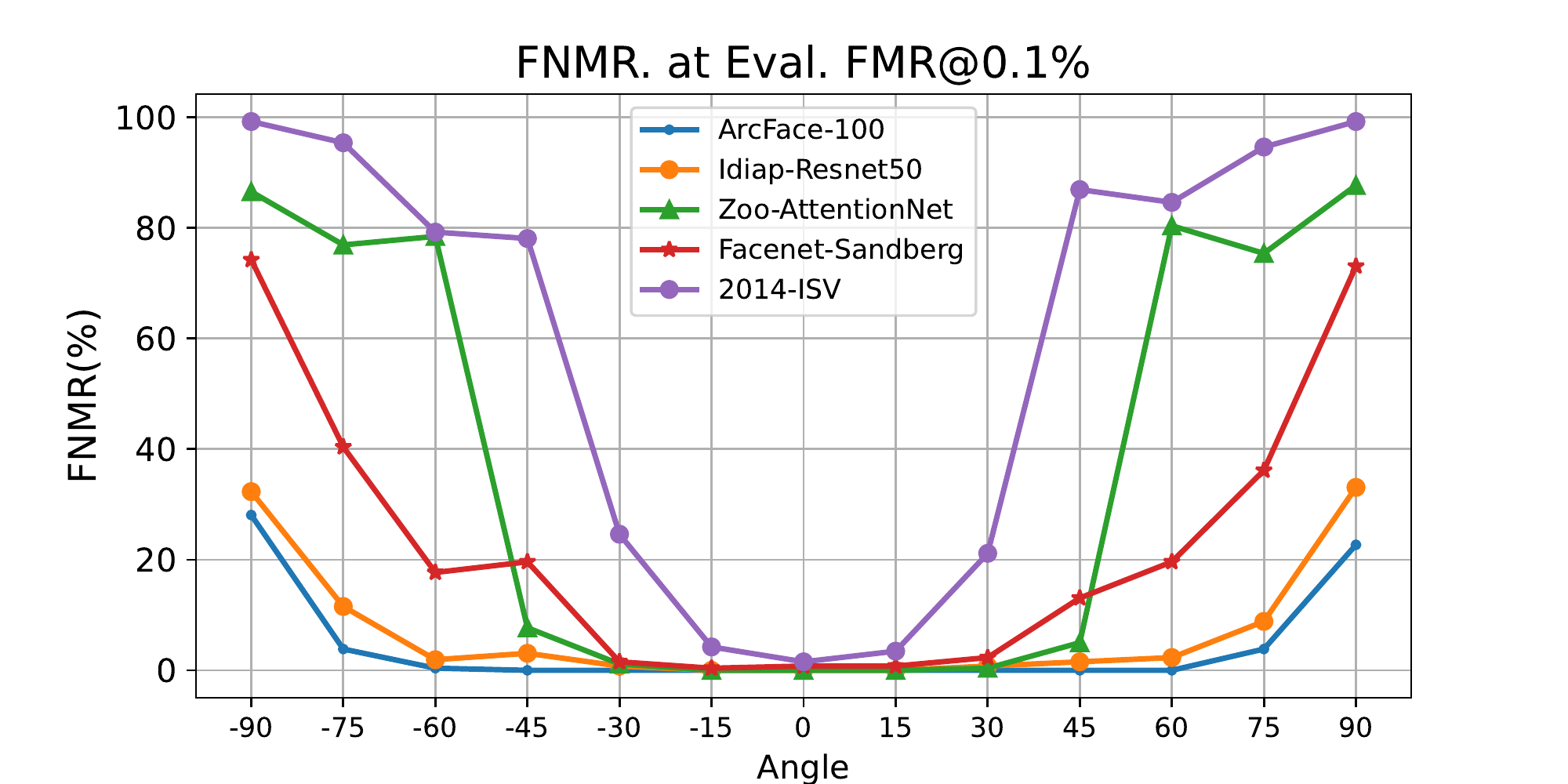}}

    \Caption[fig:thresholds]{Different Ways to Select a Threshold}{
        We depict the effect of two different wrong ways to select a score threshold.
        In \subref*{fig:thresholds:optimal}, separate score thresholds are estimated on the \emph{dev} set for each sub-protocol.
        In \subref*{fig:thresholds:eval}, separate score thresholds are selected directly on the \emph{eval} set.
        A comparison to our proposed evaluation procedure in \sfig{poses} reveals that both ways result in an unwarrented decrease in error rates.
    }
\end{figure*}

The goal of this work is threefold.
The first aim of this paper is to show the increase of face recognition performance in the last eight years, i.e., with the advent of the deep learning technology.
In our experiments we have utilized four different pre-trained deep networks that were developed at various times during the eight years, and compared their performance with the best-performing technique from our previous study \cite{gunther2016, gunther2017}.
In most of our experiments, we find that the performance decreased when the age of the algorithm increased.
For example, \textbf{2014-ISV} generally performs worst, followed by the \textbf{FaceNet-Sandberg} model developed in 2015 and the \textbf{Idiap-Resnet-50} trained in 2020 ranging third.
The second-best of the evaluated methods is the \textbf{Zoo-AttentionNet} and our winner \textbf{ArcFace-100} shows the most stable performance across all of our experiments.
While our selection covers many modern deep learning algorithms, it cannot be excluded that other inventions perform better in one or the other task.
Fortunately, our open-source, reproducible and easily extensible framework\footref{fn:package} allows to incorporate novel algorithms quickly and produces fair evaluations with respect to our tested models.

The second goal of this work is to assess what are the next steps that research should put a focus on.
We evaluate various different aspects of face recognition.
We are surprised by our finding that -- when evaluated separately -- partial occlusions and facial expressions are practically solved by our winner network \textbf{ArcFace-100} since nothing in the training procedure particularly focused on solving occlusion or expressions.
Only the datasets used for training these models include some images with expressions and mild occlusions.
For the aspect of face pose, the algorithms have improved drastically over the years.
While \textbf{2014-ISV} reaches more than 40\,\% FNMR already with angles of $\pm30\degree$, the evaluated networks can work with angles of $\pm45\degree$ or even up to $\pm75\degree$ for \textbf{Idiap-Resnet50} reasonably well; only for full-profile images, error rates are beyond expectation.
The most critical evaluation is on the SCface dataset, where none of the algorithms is able to work with the lowest-resolution faces.
Additionally, we show that open-set recognition is far from being solved and, hence, the research needs to focus more on this aspect so that the technology can be utilized for the very important task of identifying offenders in surveillance cameras.

The final goal of this paper is to show that the evaluation procedure must be adapted from how it is currently performed.
First, many evaluation protocols require to provide separate results for each protocol, and we have done the same mistake in our previous evaluation \cite{gunther2016,gunther2017}.
For example, the Celebrities in Frontal-Profile (CFP) dataset \cite{sengupta2016} provides two protocols, a frontal-frontal and a frontal-profile protocol, and more than 94\,\% accuracy is reported in \cite{mei2020}.
When evaluating both protocols separately, a different score threshold is selected for both protocols.
This is, however, not how face recognition works in practice where a single threshold is used independently of the type of image (frontal or profile) at hand.
To highlight the difference, we repeat the evaluation of the different poses from \sfig{poses} by selecting a separate threshold per sub-protocol (one threshold for each face pose) on the \emph{dev} set and plotting the FNRM on the \emph{eval} set in \sfig{thresholds:optimal}.
Clearly, the error rates drop drastically when a separate threshold is computed.
Second, in most datasets and evaluation protocols, ROC curves are plotted that show the performance of the evaluated system only on the \emph{eval} set (aka. the test set).
How well a threshold selected on this test set translates to previously unseen subjects is not clear, but from comparing \sfig{thresholds:eval} with \sfig{thresholds:optimal} we can see that there is a trend to reduce error rates when selecting the threshold on the \emph{eval} set directly.
We believe, splitting the protocols into \emph{dev} and \emph{eval} is critical to evaluate the algorithm on data that has not been seen at any stage of the process.

Finally, we want to highlight the utmost importance of reproducible research \cite{bob2017} and the requirement of providing all required details both in the paper and in code.
For example, the alignment of faces is an important step for the ArcFace network, but neither the paper nor the source code clearly shows how to do a good alignment.
Especially the alignment procedure required for handling profile images is nowhere to be found and, consequently, we had to come up with our own alignment procedure, cf.~\tab{preprocessing}, that seemed to have provided good results \cite{schmidli2021bachelor,wartmann2021bachelor}.
Only for the networks for which we know the exact alignment of profile faces (\textbf{Idiap-Resnet-50} and \textbf{FaceNet-Sandberg}), results do not abruptly degrade between $\pm45\degree$ and $\pm60\degree$.

%% file: conclusion.tex
\section{Conclusion} \label{Conclusion}
This work provides an overview on the challenges that still remains in face recognition research by running a similar set of evaluations that we carried out eight years ago in our previous work \cite{gunther2016} in a reproducible and open-source manner.
Our evaluation protocols allow an isolated examination of single aspects of face recognition (e.g., pose, occlusion, illumination, low resolution, unconstrained open-set identification), as well as a more application-oriented evaluation.
Below follow the problems that are solved well in face recognition research and the remaining challenges.

\subsection{Problems Mastered in Eight Years}

In general, we could observe that certain types of occlusion are handled well by the state-of-the-art networks using the AR face dataset as a proxy.
Two networks (\textbf{ArcFace-100}, \textbf{Zoo-AttentionNet}) presented a FNMR of 0\% at FMR 0.1\%, which is a substantial improvement from the best system we have in our previous work.
A similar trend is observed with face expressions and face recognition in mobile phones using the MultiPIE and the MOBIO datasets as respective proxies.
Illumination from different directions is no longer an obstacle, but different illumination types still constitute a gap for further research \cite{schmidli2021bachelor}.
Nevertheless, for face recognition on mobile phones, we were able to decrease FNMR from 44\% to 0\%.

\subsection{Problems Remaining to be Solved}

Despite substantial improvement, we could observe that recognition under strong pose variations is still a problem in face recognition.
Recognition under angles until 60\degree{} is very well handled by most networks.
However, once this angle increases, the number of false non-matches substantially increases as we could observe using Multi-PIE as proxy.
Recognition at a distance or with low-resolution or low-quality probe images is also an open problem in face recognition.
We could observe extremely high FNMRs using the ``Ugly'' protocol on GBU and the ``far'' protocol from SCface.
For instance, the state-of-the-art \textbf{ArcFace-100} presented an FNMR of 69\% on SCface, which is impractical in the real world.
Another open problem is open-set face recognition.
Using IJB-C as a proxy, we could observe a closed-set recognition rate of 90\% (for FPIR=100\%) using the state-of-the-art \textbf{ArcFace-100}.
This figure of merit goes down to around 35\% for a more realistic value of FPIR=0.1\%.
Finally, the reproducibility of research still is a problem.
For example, the developers of ArcFace decided to change their alignment procedure and their popular previously trained networks -- one of which we have used in our experiments -- are no longer to be found online.
Also, the alignment procedure required for profile faces in ArcFace and AttentionNet is not clear.
While we spent some effort to find optimal alignments for ArcFace both in their code and empirically \cite{schmidli2021bachelor}, the results of AttentionNet should be able to be improved with better alignment.\footnote{Some code found in the AttentionNet repository suggests that ArcFace and AttentionNet use the same image alignment, but it is unclear if this code was actually used to align their images.}
We are utilizing the open-source and reproducible face recognition framework of Bob \cite{guenther2012facereclib} and providing all relevant details of all our experiments.\footref{fn:package}
This makes our research distinct from other reviews that can only rely on results reported in the literature since they cannot re-run experiments or change evaluation metrics.

\vspace{1em}

While some problems still remain to be solved, we could observe great progress in face recognition research in the last eight years.
It is worth noting that none of the tested algorithms were carefully crafted to handle the above-mentioned aspects.
The availability of large amounts of data definitely plays an important role in the recent state-of-the-art networks.
Furthermore, the three best systems we presented use different variations of the ArcFace loss \cite{deng2018arcface}, which definitely played an important role as well.

We are aware that we only used a small subset of available deep networks for face recognition, and we are sorry if we missed your particular network.
Furthermore, this work did not consider security aspects in face recognition systems, such as morphing or presentation attacks \cite{scherhag2019face,marcel2019handbook}.
Possible extensibility to cover these aspects would require new work with a new experimental setup that we were not able to cover in this one.
Fortunately, the source code for this study is publicly available,\footref{fn:package} and new implementations \cite{linghu_zhang2021master_project} in Bob's biometric recognition framework \cite{bob2012,bob2017,guenther2012facereclib} allow for a very easy extension to include your network.